\def\year{2022}\relax
\documentclass[letterpaper]{article} 
\usepackage{aaai22}  
\usepackage{times}  
\usepackage{helvet}  
\usepackage{courier}  
\usepackage[hyphens]{url}  
\usepackage{graphicx} 
\usepackage{natbib}  
\usepackage{caption} 
\DeclareCaptionStyle{ruled}{labelfont=normalfont,labelsep=colon,strut=off} 
\usepackage{algorithm}
\usepackage{algorithmic}

\usepackage{newfloat}
\usepackage{listings}
\floatstyle{ruled}
\newfloat{listing}{tb}{lst}{}
\floatname{listing}{Listing}
\usepackage{mwe}
\usepackage{amsmath}
\usepackage{amssymb}
\usepackage{mathtools}
\usepackage{bbm}
\usepackage{subcaption}

\title{Polynomial-Spline Neural Networks with Exact Integrals}
\author{
    Jonas A. Actor,\textsuperscript{\rm 1} Andy Huang,\textsuperscript{\rm 2} Nathaniel Trask\textsuperscript{\rm 1}
}
\affiliations{
    Affiliations
    \textsuperscript{\rm 1}Center for Computing Research, Sandia National Laboratories\\
    \textsuperscript{\rm 2}Radiation and Electrical Science, Sandia National Laboratories \\
    1515 Eubank Blvd SE \\
    Albuquerque, NM 87123 \\
    jaactor@sandia.gov, ahuang@sandia.gov, natrask@sandia.gov
}

\begin{document}

\maketitle

\begin{abstract}
Using neural networks to solve variational problems, and other scientific machine learning tasks, has been limited by a lack of consistency and an inability to exactly integrate expressions involving neural network architectures.
We address these limitations by formulating a novel neural network architecture incorporating free knot B1-spline basis functions into a polynomial mixture-of-experts model. Effectively, our architecture performs piecewise polynomial approximation on each cell of a trainable partition of unity while ensuring the neural network and its derivatives can be integrated exactly, obviating a reliance on sampling or quadrature and enabling error-free computation of variational forms. We demonstrate $hp$-convergence for regression problems at convergence rates expected from approximation theory and solve elliptic problems in one and two dimensions to show an effective dimension reduction compared to adaptive finite elements.

\end{abstract}

\section{Introduction}
\noindent 
Deep neural networks (DNNs) have been proposed for solving partial differential equations (PDEs) and scientific machine learning, but fail to converge to PDE solutions in practical settings \cite{dnnerror, pdeDL}. 
Notably, partition of unity networks (POUNets) \cite{pounet} demonstrate $hp-$ convergence for regression problems, in terms of spatial resolution $h$ and polynomial degree $p$, similar to finite element methods.
However, use of inexact quadrature is a \textit{variational crime} \cite{strang1972variational} complicating error analysis for these and other architectures; such issues have led to the popularity of simpler to implement but more difficult to analyze collocation schemes \cite{pinns,pinnerror}.

In this work, we introduce polynomial-spline neural networks, a mixture-of-experts (MOE) model that combines gating functions composed of convex combinations of B-spline basis functions, with polynomial experts localized to each cell of the partition.
The spline gating functions lead to a piecewise polynomial approximation with explicitly parameterized support, creating closed-form expressions for the integral of the DNN and its derivatives. As such, this provides a foundation for other problems requiring integration, e.g. estimation of statistical moments for probability measures, or novel loss functions and regularizers.

\section{Related Work}

Our architecture admits interpretation as a MOE model \cite{mixtureofexperts} with gating functions provided by a convex combination of B-spline basis functions rather than a hidden layer with softmax activation. While similar to MOE, POUNets \cite{pounet} aim to build $hp-$convergent solutions to regression/PDE problems rather than perform kernel density estimation. Unlike the multilayer preceptron (MLP) or radial basis function network considered in \cite{pounet}, working with convex combinations of B-splines admits closed form expressions for integrals in terms of the B-spline knots. We thus preserve the $hp-$ convergence of POUNets while enabling exact integration, even in high-dimensional settings; Compared to traditional B-spline bases the convex combination identifies an optimal compressed subspace for nonlinear approximation.

Our B-splines provide natural extensions of well-known results interpreting ReLU networks as continuous piecewise linear (CPWL) functions \cite{reluLP}. Previous works use splines to study approximation properties of DNNs, treating e.g.  ReLU networks as max-affine splines \cite{maso} or as P1 finite elements \cite{relufem}. In high-dimensions, the CPWL interpretation of ReLU networks is not tractable for quadrature, as the geometrically complex piecewise linear regions form non-convex polyhedral domains and do not admit a closed-form description of their support. In light of these results, training a ReLU DNN is similar in scope to finding an optimal spline interpolant with trainable knots \cite{maso}, an NP-hard problem \cite{freespline}. In the related problem of adaptively finding an optimal mesh for discretizing PDEs, the mesh adaptivity is guided by an energy functional relating to the discretization points, whose minimizer is the optimal adaptive mesh. Such an energy function again necessitates integrating over the variational form of the PDE \cite{adaptive1d, adaptivehp, fenicsbook}, which precludes their use in training DNNs.

Some PDE discretizations of DNNs skirt the issue of integration by adapting a collocation PDE residual (e.g.
physics-informed neural networks (PINNs) \cite{pinns} and related methods). While effective, this requires strong regularity requirements and more involved mathematical analysis beyond the standard Lax-Milgram theory \cite{shin2020convergence}. Alternatively, the Deep Ritz method uses as a loss the Euler-Lagrange functional of the relevant variational problem, but ultimately resorts to sampling-based methods for integration \cite{deepritz}. As a result, the convergence of the loss function is dominated by the error in Monte Carlo integration, and variational crimes complicate the already complex landscape of approximation error, optimization error, and stability theory. An important practical feature of preserving the variational form is that we may train by evolving along the manifold of optimal fits to data, similar to the least-squares gradient descent optimizer \cite{lsgd}.

\section{Formulation}

Let $\Omega \subset \mathbb{R}^d$ be a closed, compact domain, where $d$ is the spatial dimension; assume for simplicity that $\Omega = [0,1]^d$. Let $\mathbb{P}^{B}(\Omega)$ be the space of polynomials of degree up to $B$ on $\Omega$, with basis $\{p_\beta\}_{\beta=1,\dots,d_P = \text{dim}(\mathbb{P}^B(\Omega))}$.

We define a \emph{polynomial-spline neural network} $y: \Omega \rightarrow \mathbb{R}$ via the expression
\begin{equation}
y(x) = \sum_{\alpha=1}^{N_{\text{cells}}} \left( \sum_{\gamma=0}^{N_{\text{splines}}} w_{\alpha, \gamma} \phi_\gamma(x) \right) \left( \sum_{\beta=1}^{d_P} c_{\alpha, \beta} p_{\beta}(x) \right). \label{eq:pou-def}
\end{equation}
In this expression, the functions $\phi_\gamma: \Omega \rightarrow \mathbb{R}$ are B-spline basis functions, parameterized by a set of knots $\{t_{\gamma}\} \subset \Omega$. Additionally the coefficients  $w_{\alpha,\gamma}$ are constrained so that for all $\gamma$, $\sum_{\alpha=1}^{N_\text{splines}} w_{\alpha, \gamma} = 1$ and $w_{\alpha, \gamma} \ge 0$. When clear, the bounds for the summations in Equation \eqref{eq:pou-def} are dropped for convenience.

The functions $\varphi_\alpha(x) = \sum_{\gamma=1}^{N_\text{splines}} w_{\alpha, \gamma} \phi_\gamma(x)$ form a partition of unity (POU) of $N_\text{cells}$ partitions, following from convexity of $W_{\alpha,\gamma}$ and the fact that B-spline basis functions form a partition of unity \cite{powell}.
Thus, polynomial-spline networks are MOE models, where convex combinations of B-splines serve as gating functinos for $B^{th}-$ order polynomial experts.
In the case that $N_\text{cells}=1$, training this architecture reduces to polynomial approximation, as the polynomial-spline network becomes \begin{equation*} \begin{split}
y(x) 
&= 1 \cdot \left( \sum_{\beta} c_{\beta} p_\beta(x) \right) \\
&= \sum_{\beta} c_{\beta} p_\beta(x).
\end{split} \end{equation*}
Similarly, in the case $B=0$ and $N_\text{cells} = N_\text{splines}$, training this architecture reduces to free-knot spline approximation, since the network becomes
\begin{equation*} \begin{split}
y(x) &= \sum_{\alpha} \left( \sum_\gamma w_{\alpha, \gamma} \phi_\gamma(x) \right) \left( c_\alpha \right) \\
&= \sum_\alpha \sum_\gamma c_\alpha w_{\alpha,\gamma} \phi_\gamma(x),
\end{split} \end{equation*}
and setting $w_{\alpha,\gamma} = \delta_{\alpha\gamma}$ reduces the architecture to
\begin{equation*} \begin{split}
y(x) &= \sum_{\alpha} c_\alpha \phi_\alpha(x).
\end{split} \end{equation*}
Therefore, we expect our polynomial-spline network to exhibit some form of both $h-$ and $p-$ refinement, as the number of partitions and polynomial degree increase, respectively, following the convergence rates established via numerical analysis e.g. \cite{sulimayers}.

In practice, we limit ourselves to only using B1-splines. In doing so, we make the max-affine spline interpretation of deep ReLU networks \cite{maso} explicit, in that we directly construct the underlying spline to partition $\Omega$. Doing so allows us to construct closed-form expressions for the integrals (and integrals of derivatives) of the polynomial-spline network; deriving such expressions is tedious but feasible for higher-order splines.

We outline how to construct analytic expressions for the integral of $y$ in the case of the functions $\phi_\gamma$ being B1-spline basis functions and our domain $\Omega = [0,1]$.  First, note that the B1-spline basis functions are described entirely by the set of knots $\{t_\gamma\}_{\gamma=0,\dots,N_{\text{splines}}}$, with $t_0 = 0$ and $t_{N_\text{splines}} = 1$, with relation to $\phi_\gamma$ in that $\phi_\gamma(t_\gamma) = 1$ and $\phi_\gamma(t_\beta) = 0$ for all $\beta \ne \gamma$. 
By construction, when restricted to the interval $[t_{\gamma-1},t_\gamma]$, the polynomial-spline network $y$ is a polynomial of degree $B+1$.
Letting $\{q_i\}_{i=1,\dots,d_P+1}$ be a basis for $\mathbb{P}^{B+1}([t_{\gamma-1}, t_\gamma])$, we express $y$ restricted to our interval in this basis, i.e.
{\small{
\begin{equation*} \begin{split}
y(x) &= \sum_\alpha \left( w_{\alpha, \gamma-1} \phi_{\gamma-1}(x) + w_{\alpha, \gamma} \phi_{\gamma}(x) \right) \left( \sum_{\beta} c_{\alpha\beta} p_{\beta}(x) \right) \\
:&= \sum_{i=1}^{d_P + 1} d_i q_i(x)
\end{split} \end{equation*}
}}
for coefficients $d_i$, which are closed-form expressions of the coefficients $w_{\alpha, \gamma}$, $t_\gamma$, and $c_{\alpha\beta}$.
Therefore, our integral becomes 
\begin{equation*} \begin{split}
\int_\Omega y(x) dx &= \sum_{\gamma=1}^{N_\text{splines}} \int_{t_{\gamma-1}}^{t_\gamma} y(x) dx \\
&= \sum_{\gamma=1}^{N_\text{splines}} \sum_{i=1}^{d_P+1} d_i \int_{t_{\gamma-1}}^{t_\gamma}  q_i(x) dx. \end{split} \end{equation*}
Choosing the monomial basis $q_i(x) = x^{i-1}$, 
\begin{equation*} \begin{split}
\int_\Omega y(x) dx
&= \sum_{\gamma=1}^{N_\text{splines}} \sum_{i=1}^{d_P+1} d_i \int_{t_{\gamma-1}}^{t_\gamma} x^{i-1} dx \\
&= \sum_{\gamma=1}^{N_\text{splines}} \sum_{i=1}^{d_P+1} \frac{d_i}{i} \left( t_{\gamma}^{i} - t_{\gamma-1}^{i} \right).
\end{split} \end{equation*}
Since we can explicitly calculate expressions for $d_i$, and all other values are known weights in our network, we can use the above formula to directly integrate $y$. Similar expressions are derived in the same way for $\nabla y$, or for the calculation of moments involving $y$ to a power, by expressing the integrand in terms of a polynomial expansion in terms of the polynomials $q_i$ and then integrating separately upon the support of each B1-spline basis function. 

\section{Experiments}

We demonstrate the effectiveness of our architecture on two sets of problems: regression problems and variational problems. Our implementation of polynomial-spline neural networks, and all the related layers and training strategies described herein, are implemented using TensorFlow \cite{tf}; we use NumPy \cite{numpy}, SciPy \cite{scipy}, and FEniCS \cite{fenics-software} to compare our results to classical methods. 

\subsection{Construction}

To build our polynomial-spline network, we build B1-spline basis functions for $\Omega$ as a tensor product of 1D B1-spline basis functions along each dimension. For each dimension, we construct a B1-spline layer, whose knots are parameterized to accommodate TensorFlow backwards differentiation during training. The general expression for a hat function built via ReLU functions is given in \cite{relufem}. 
During training however, knots may become unordered, leading to a problematic inversion of elements. To prevent this concern, we track the relative position between knots rather than the locations themselves, constraining them to span the extant $\Omega$. For more details, please see the supplementary material.

\subsection{Expedited Training via LSGD}

We expedite the training of our models by using the least-squares gradient descent optimizer(LSGD)  \cite{lsgd}. We define the function $\Phi: \mathbb{R}^d \rightarrow \mathbb{R}^{N_\text{cells} d_P}$ as $\Phi_{\alpha\beta}: x \mapsto \left( \sum_\gamma w_{\alpha,\gamma} \phi_\gamma(x) \right) p_\beta(x)$, and rewrite Equation \ref{eq:pou-def} as $$ y(x) = c^T \Phi(x)$$ for a vector of coefficients $c \in \mathbb{R}^{N_\text{cells} d_P}.$ For regression problems, the LSGD solver adds a least-squares solve for $c$ between each gradient step of the first-order optimizer. We wrap this least-squares solve call into a custom TensorFlow layer so that the operation is embedded in the network's TensorFlow graph directly, enforcing that all outputs of the network lie on the manifold of best-fit solutions regarding the coefficients of the outermost layer.
We note that there are stability issues with the implementation of the backwards differentiation of the least-squares function call in TensorFlow, resulting in accuracies at most of order $O(10^{-8})$ instead of machine-epsilon precision. Other implementations of this method do not suffer from this issue and can achieve errors as low as $O(10^{-20})$ and beyond; see the supplementary material for more details.

For regression,  LSGD  solves the least-squares problem, given batch data $\{x_i, y_i\}_{i=1,\dots,\text{batch size}}$ the problem
$$ \min_c \left \lVert y - c^T \Phi(x) \right \rVert_F^2.$$
For variational problems, the least-squares problem that we solve is the variational problem that corresponds to the Euler-Lagrange functional; see the discussion below about our variational problems for more details.


\subsection{Training Details}
For the regression problems, we train on a random uniform set of 1000 points, and we validate our model on a separate random uniform set of the same size. For the variational problems, there are no data sites involved, with the loss defined via the closed form expression for the energy. For all problems we use the Adam optimizer \cite{adam} . Our loss for the regression problems is mean squared error (MSE), while for the variational problems our loss is the Euler-Lagrange functional i.e. the Ritz energy. More details, including a description of hyperparameters, are available in the supplementary material.
Code for each problem is available in the supplementary material. All code is run on a 64-core Intel Xeon Gold CPU running Linux with 10 NVIDIA Tesla V100 GPUs with 32GB each, although code was restricted at runtime to only use 1 GPU and 10GB of memory for portability reasons.

\subsection{Problems}
\label{sec:results}
To demonstrate the effectiveness of our architecture, we pose two sets of problems. First, we deploy our architecture on regression problems to evaluate consistency. After, we progress to solving variational problems.

\subsubsection{Regression}
We test our architecture on two 1D regression problems:
\begin{enumerate}
\item $f(x) = \sin( 2 \pi x)$ for $x \in [0,1]$ \label{reg:hp}
\item $f(x) = \left \lvert \sin(3 \pi x^2) \right \rvert + \left \lvert \cos(5 \pi x^2) \right \vert$  for $x \in [0,1]$   \label{reg:adaptive}
\end{enumerate}
In Problem \ref{reg:hp}, we expect to see both $h-$ and $p-$ refinement, i.e. increasing the number of partitions or increasing the degree of polynomial approximation, respectively, should improve the approximation. In Problem \ref{reg:adaptive}, we expect $h-$ refinement to improve our approximation but $p-$ refinement to not, since the function $f$ in this case is only piecewise smooth. However, if we have a sufficient number of knots (i.e. at least one per piece of $f$), our POU cells should adaptively during training recover the locations where the $\sin$ or $\cos$ terms change sign, and that on each piece, we expect $p-$ refinement to improve our approximation.

\paragraph{Problem \ref{reg:hp}.}

We perform three experiments to demonstrate the $hp-$ refinement properties of our network. First, we fix $N_\text{cells}=1$ to verify that our model compares favorably to polynomial approximation with regards to $p-$ refinement. Second, we fix the polynomial degree $B = 0$ to verify that our model compares favorably to spline approximation with regards to $h-$ refinement. Third, we test our model for a variety of parameters for polynomial degree, number of cells, and number of spline knots, to verify simultaneous $hp-$ refinement and to test that our model can capture the solution to this regression problem up to machine precision.

\begin{figure}[t]
\centering
\includegraphics[width=0.9\columnwidth]{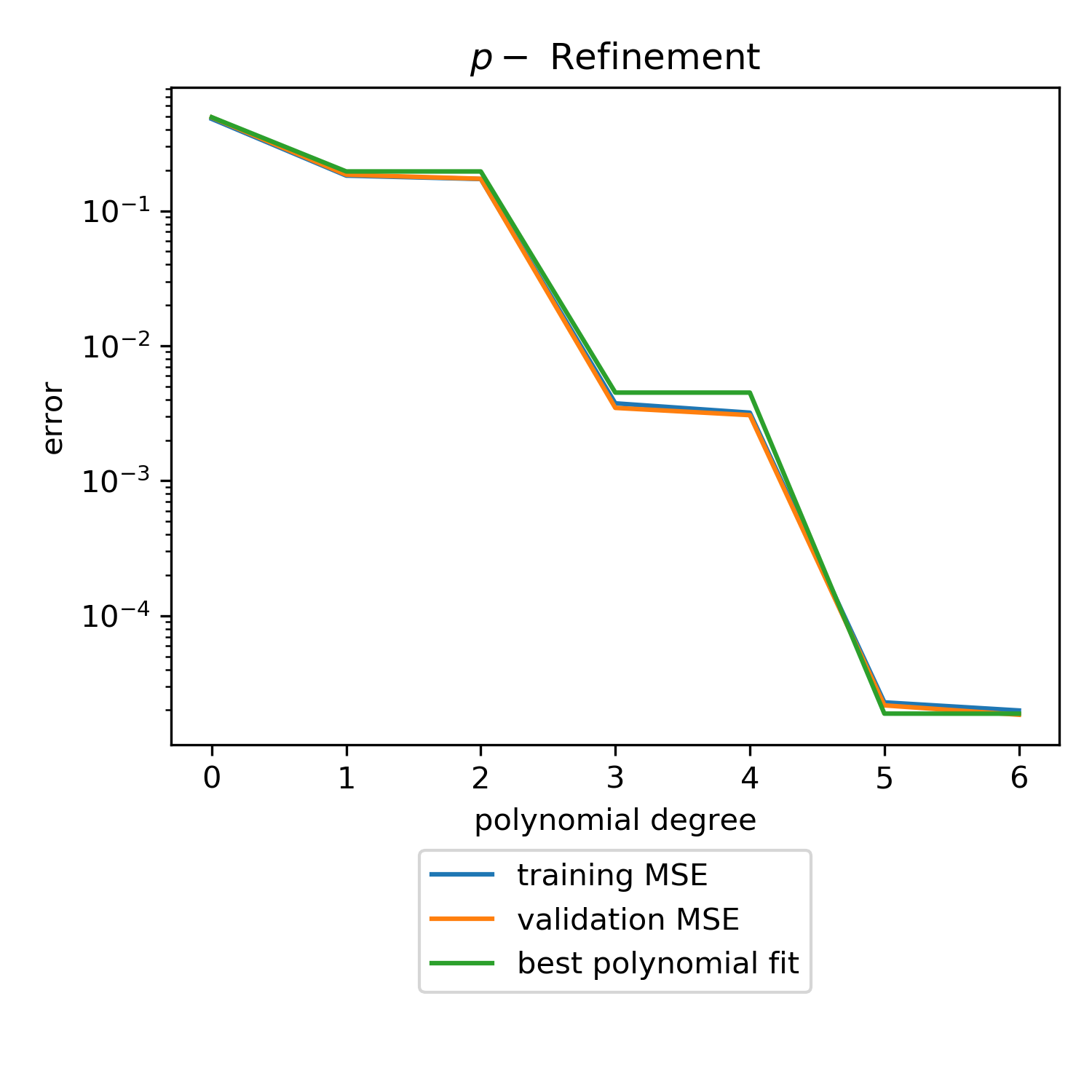}
\caption{ Results of $p-$ refinement for regression Problem \ref{reg:hp}, with $N_\text{spline}=4$ and $N_{\text{cells}}=1$. The errors for the training and validation sets coincide with the error for the best polynomial approximation. \label{fig:reg1-p}}
\end{figure}

First, we consider Problem \ref{reg:hp} with $N_{\text{cells}}=1$; in this case, we expect our model to return the best polynomial approximation of the specified degree. Results are shown in Figure \ref{fig:reg1-p}; all three lines plotted in the figure nearly coincide, showing that in this limit our model maintains the $p-$ refinement properties of polynomial approximation. The staircase phenomenon is due to the function $f(x) = \sin(2 \pi x)$ being an odd function, and as such the best polynomial approximation (and our model) only improves when the polynomial degree is increased to an odd power.

\begin{figure}[t]
\centering
\includegraphics[width=0.9\columnwidth]{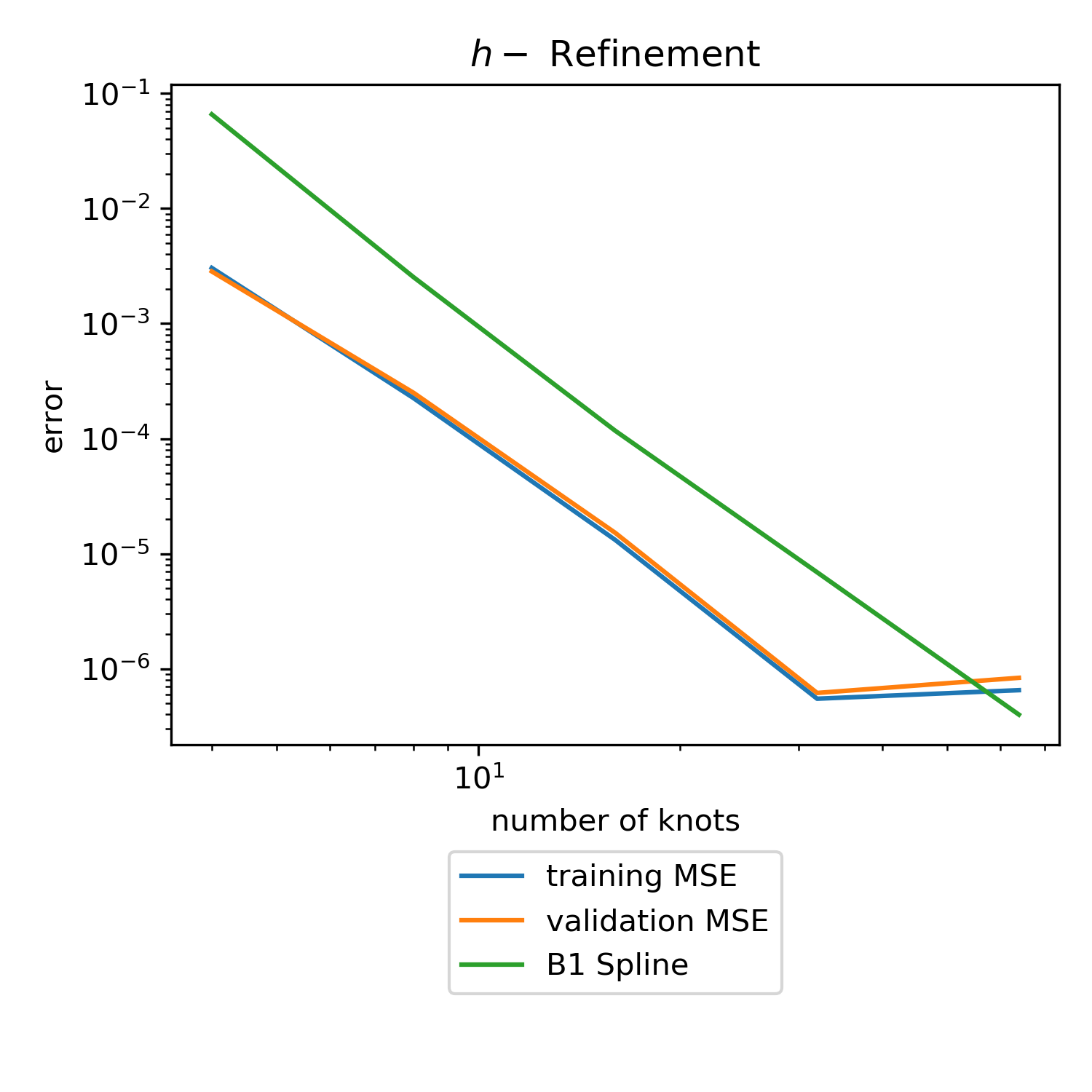}
\caption{ Results of $h-$ refinement for regression Problem \ref{reg:hp}, with $N_{\text{cells}} = N_{\text{spline}}$ and polynomial degree $B = 0$. The $x$-axis maintains a logarithmic scale. \label{fig:reg1-h}}
\end{figure}

Second, to verify the $h-$ refinement properties of our architecture, we set the polynomial degree $B = 0$ and compare our results to a piecewise linear spline approximation with uniform knots. Error plots are shown in Figure \ref{fig:reg1-h}. In this case, we see the roughly the same rate at which the error decreases as we increase the number of knots in the network and in our spline approximation, until our model plateaus due to the instability of the backwards differentiation in the LSGD layer. 

\begin{figure}[t]
\centering
\includegraphics[width=0.9\columnwidth]{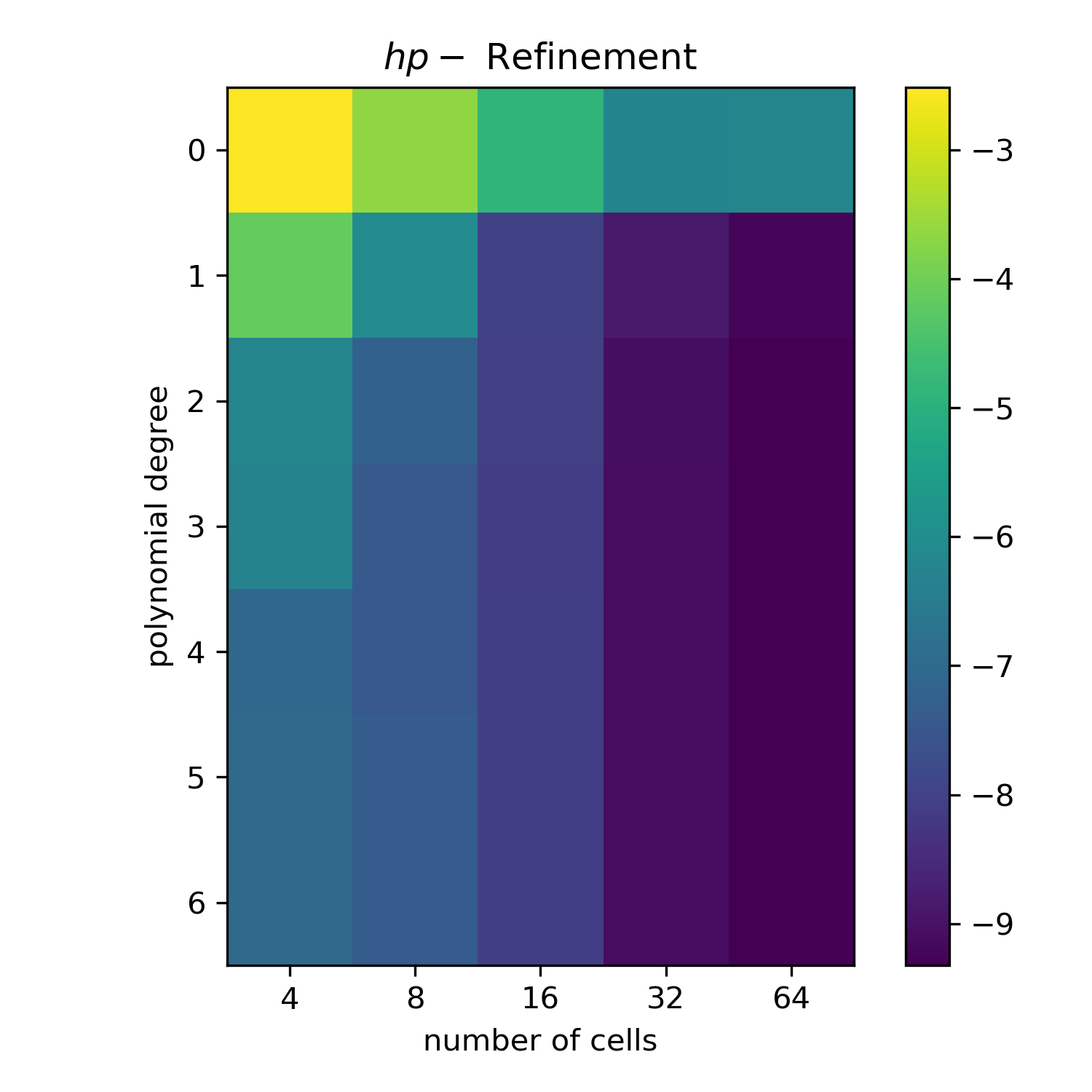}
\caption{Results for $hp-$ refinement on regression Problem \ref{reg:hp}. The color bar maps the $\log_{10}$ of the mean squared error for the specified number of POU cells and polynomial degree. \label{fig:reg1-hp}}
\end{figure}

Finally, when using $h-$ and $p-$ refinement simultaneously the polynomial-spline network is capable of achieving machine precision accuracy. Results are shown in Figure \ref{fig:reg1-hp}.
We see that for sufficient spatial resolution and polynomial degree, we achieve mean-squared errors of order $10^{-8}$ or better using our proposed network, which is the precision limit given by our least-squares implementation in the LSGD layer.

\paragraph{Problem \ref{reg:adaptive}.}

We perform two sets of experiments for Problem \ref{reg:adaptive}. First, we perform the same experiment as for Problem \ref{reg:hp} to test $h-$ refinement even in the case of a non-smooth target function. Second, we test whether our network can find the discontinuities in the derivative of $f$ in a way that is comparable to piecewise polynomial approximation. As in the previous problem, the training and validation error curves are extremely similar, so we only plot validation errors in the rest of the figures in this section.

First, as before, we compare our network's performance when the polynomial degree $B=0$ and compare to a spline approximation with the same number of knots.
\begin{figure}[t]
\centering
\includegraphics[width=0.9\columnwidth]{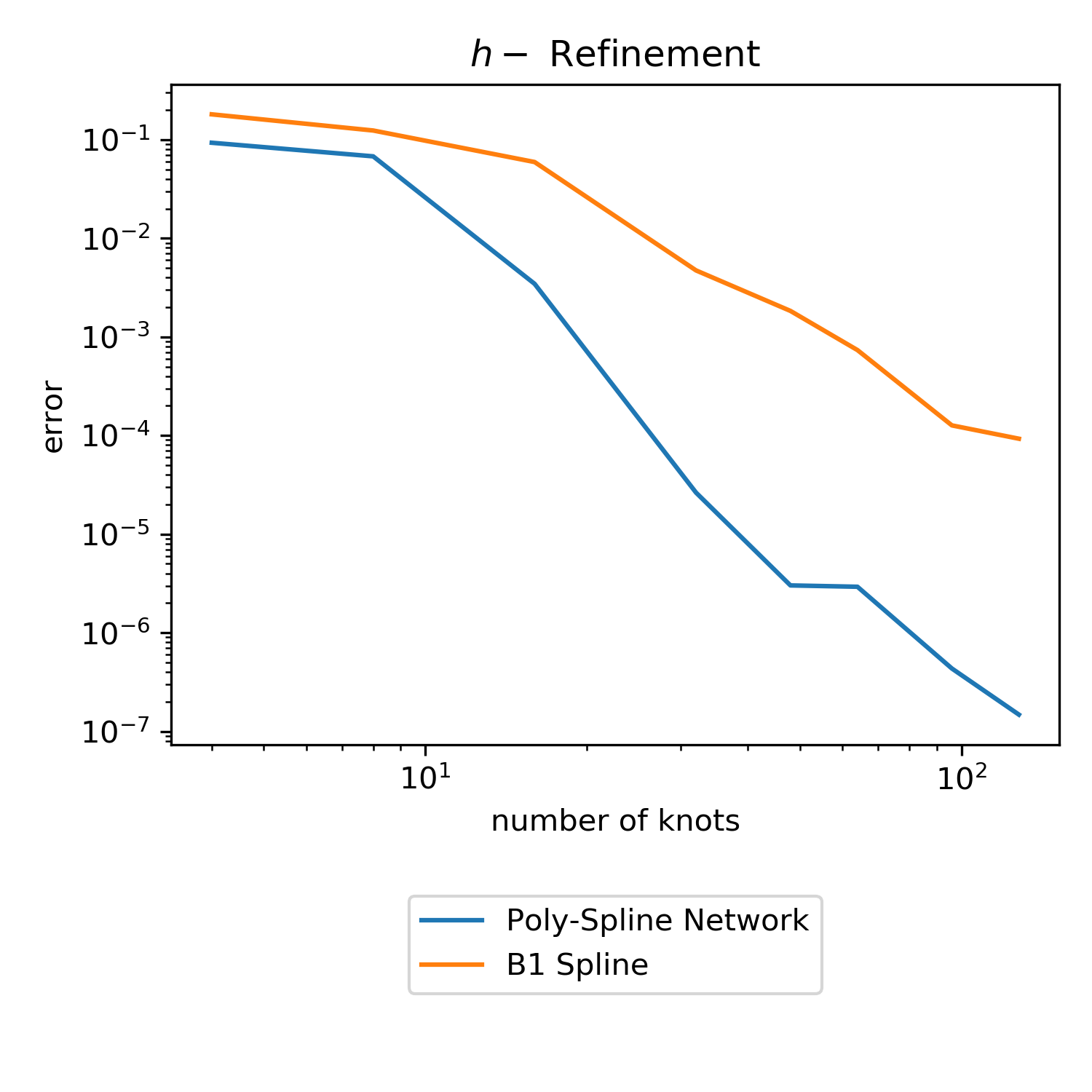}
\caption{Training results for Problem \ref{reg:adaptive} with $B=0$ i.e. spline approximation, with $N_\text{spline} = N_\text{cells}$ plotted against the achieved mean squared error.  \label{fig:reg2-h}}
\end{figure}
In Figure \ref{fig:reg2-h}, we see the error in our approximations for increasingly larger numbers of knots; we observe a roughly log-linear decrease in mean squared error as we double the number of knots (and POU basis functions) in our network, as expected.

\begin{figure}[t]
\centering
\includegraphics[width=0.9\columnwidth]{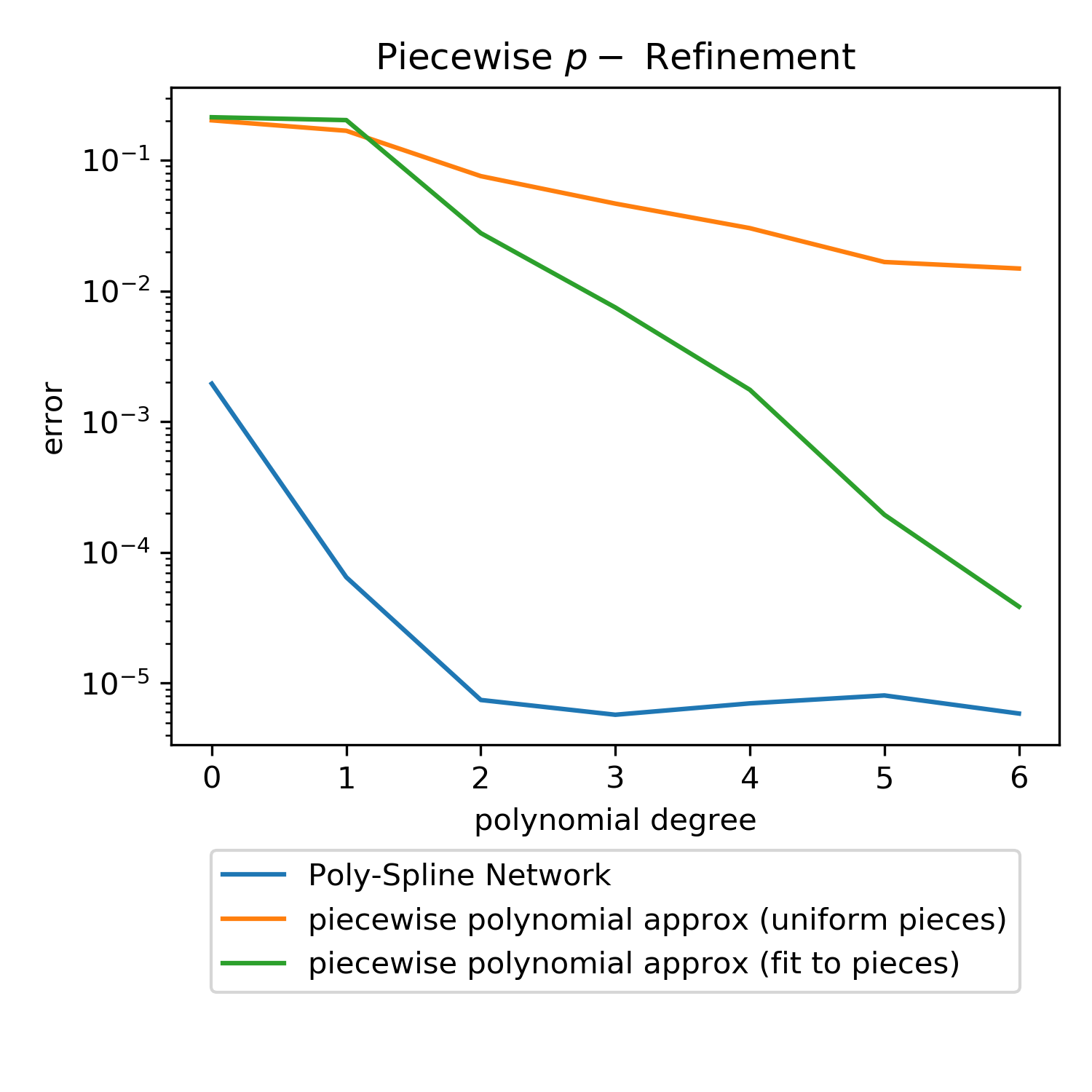}
\caption{ Results of piecewise $p-$ refinement for Problem \ref{reg:adaptive}, with $N_\text{cells}=8$. \label{fig:reg2-adaptive}}
\end{figure}

Second, we compare the results of our model with comparable piecewise polynomial approximation problems. We specifically compare to two piecewise polynomial approximations, one where polynomials are fit upon uniform pieces, and the other fit to pieces where the derivative of $f$ is discontinuous. As both of these piecewise polynomial approximants fit a total of 8 polynomials, we compare these results to our model with $N_\text{cells} = 8$, which fits 8 polynomials in the LSGD layer. Results are shown in Figure \ref{fig:reg2-adaptive}.
Our polynomial-spline network outperforms these model for all polynomial degrees, efficiently capturing the discontinuous derivatives in the target function. The success of the polynomial-spline network plateaus at an accuracy of $O(10^{-6})$ due to the limitations of the numerical stability in the automatic differentiation of the least-squares solve operation in the LSGD layer.

\subsubsection{Variational Problems}
We test our architecture on two variational problems:
\begin{enumerate}
\setcounter{enumi}{2}
\item \label{var:1d} 1D Poisson problem with Dirichlet boundary conditions:
\begin{equation}
\begin{array}{rll}
- d^2u = 2 & \text{ on } & \Omega = [0,1] \\
u = 0 & \text{ at } & \partial \Omega = \{0,1\}.
\end{array} \end{equation}
\item \label{var:slit} 2D Poisson problem on a slit domain:
\begin{equation}
\begin{array}{rllll}
- \Delta u & = 0 & \text{ on } & \Omega = [-1,1]^2 \\
u  &= g(r,\theta) & \text{ on } & \Gamma = \partial \Omega \cup \,\, [0,1] \times \{0\},
\end{array} \end{equation}
where $g(r,\theta) = \sqrt{r} \sin\left(\frac{\theta}{2} \right)$ is given in polar coordinates.
\end{enumerate}
In Problem \ref{var:1d}, we expect to recover the true solution $u(x) = x(1-x)$ exactly (up to machine precision), assuming on each POU cell we are fitting a polynomial of sufficient degree. In Problem \ref{var:slit} we expect our problem to recover the known singularity at the origin similar to adaptive mesh refinement methods \cite{adaptive1d, adaptivehp, fenicsbook}; the true solution to this problem is known to be $u = g(r,\theta)$.

For each of these variational problems, we minimize the associated  Euler-Lagrange functional \cite{evans}. 
For example, Problem \ref{var:1d} is solved by minimizing the Euler-Lagrangian ``loss''
\begin{equation} 
L(u) = \int_\Omega \frac{1}{2} \left \lVert \nabla u \right \rVert^2 - 2 u + \beta \left( u(0)^2 + u(1)^2 \right),
\label{eq:EL-var1}
\end{equation}
where $\beta > 0$ is a penalty parameter to enforce our solutions satisfy our boundary conditions. For Problem \ref{var:slit}, the Euler-Lagrange functional is
\begin{equation}
L(u) = \int_\Omega \frac{1}{2} \left \lVert \nabla u \right \rVert^2 + \beta \int_{\Gamma} (u - g(r,\theta))^2.
\label{eq:EL-var2}
\end{equation}

Instead of relying on Monte Carlo or sampling-based methods to evaluate these integral (e.g. \cite{deepritz}), we compute these integrals exactly by virtue of our architecture construction. Rather than using the analytic expressions for the integral, we employ Gaussian quadrature of degree $B + d + 1$, which computes the integrals exactly for polynomials of degree up to $2(B + d) + 1$, which is sufficient for exactly computing the integrals of $u$, $u^2$, and $\left \lVert \nabla u \right \rVert^2$ on the support of each B1 basis function; using quadrature allows us to exploit hardware acceleration during computation, since the model evaluation at the quadrature points can more efficiently use the GPUs. We note that since the location of the knots change over the course of training, the quadrature points change as well.

\paragraph{Problem 3: 1D Poisson Boundary Value Problem}

For this problem, the LSGD layer solves the linear system corresponding to the weak form of Problem \ref{var:1d}, which is given as follows. 
We define the matrix $A$ and vector $b$ by substituting $u(x) = c^T \Phi(x)$ into Equation \ref{eq:EL-var1}, taking the derivative of Equation \ref{eq:EL-var1} with respect to $c$, and then setting the resulting expression to zero. In doing so, we arrive at the linear problem $Ac = b$, where
\begin{equation*} \begin{split}
A &= \int_\Omega d\Phi d\Phi^T + \beta\left( \Phi(0) \Phi(0)^T + \Phi(1) \Phi(1)^T \right) \\
b &= \int_\Omega 2 \Phi.
\end{split} \end{equation*}
The integrals in the linear problem can be calculated exactly for a given set of knots, since upon each component of the B1-spline functions, $\Phi(x)$ is a polynomial of degree $B+d$ i.e. $B+1$. We can do so by deriving the closed-form solutions for the integrands of each interval between spline knots, or (more efficiently) by using Gaussian quadrature of degree $B+d+1$ as specified earlier.

\begin{figure*}[t]
\centering
\includegraphics[width=0.8\textwidth]{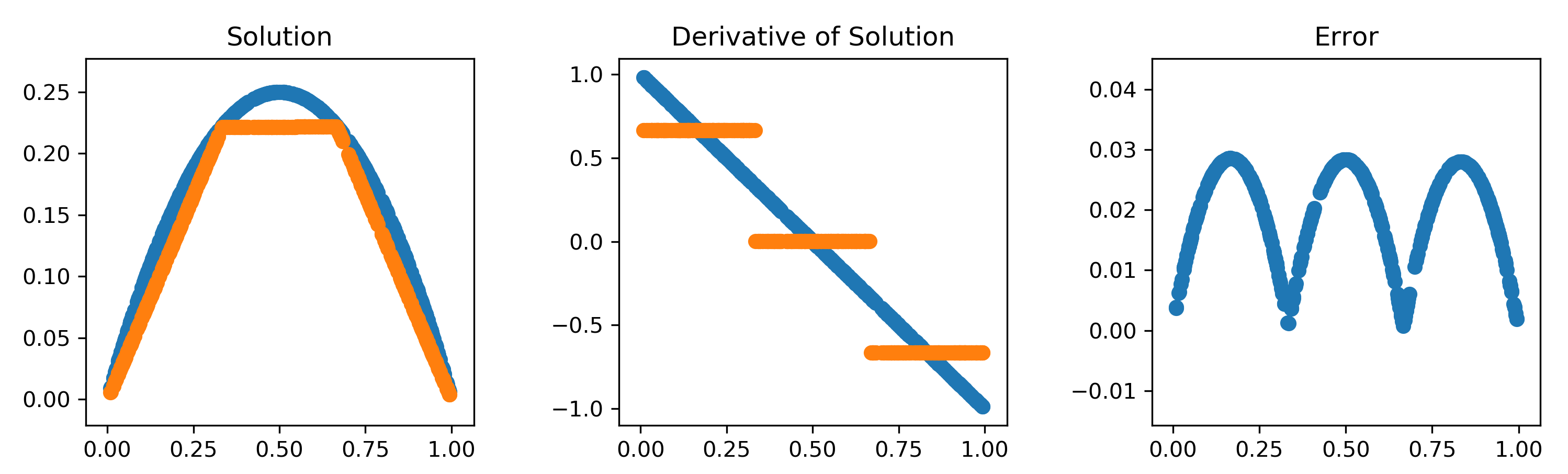}
\includegraphics[width=0.8\textwidth]{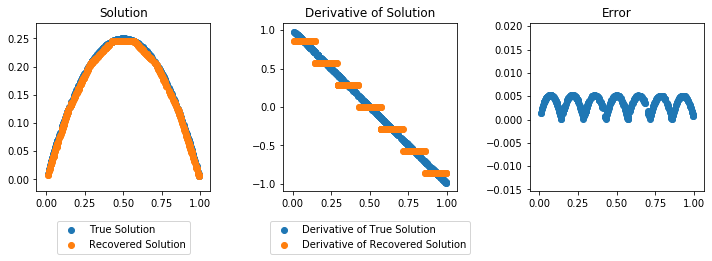}
\caption{ Recovered solution to Problem \ref{var:1d} with $N_\text{cells}=3$, $N_\text{spline}=4$, and $B=0$ (top) and with $N_\text{cells}=3$, $N_\text{spline}=8$, and $B=0$ (bottom). We recover the best FEM approximation on a mesh with $N_\text{spline}$ nodes i.e. $N_\text{spline} -1$ intervals, as seen in the plot of the derivative of the approximation in the center panel. \label{fig:var1-4}}
\end{figure*}

When we use the training methods described earlier, we successfully solve our problem up to machine precision (or the limits of the accuracy of the least-squares backwards differentiability) nearly immediately when $B\ge 1$, which matches our expectation, since the true solution to our problem is a polynomial of degree 2. Please refer to the supplementary material for more details on these results. In the case $B=0$, we recover the best P1-finite element solution to our problem, as seen in Figure \ref{fig:var1-4}, along with the expected effects of mesh refinement, as seen by the difference in the error plots between the two figures.

\paragraph{Problem 4: 2D Poisson Slit Domain}

By symmetry, it suffices to solve this problem on a reduced domain $\Omega'$ that is half the size of $\Omega$: note that the solution to Problem \ref{var:slit} is symmetric about the x-axis. As a result, we solve the following equivalent problem:
\begin{equation}
\begin{array}{rllll}
- \Delta u & = 0 & \text{ on } & \Omega' = [-1,1] \times [0,1] \\
\partial_n u & = 0 & \text{ on } & \Gamma_N = [-1,0] \times \{0\} \\
u  &= g(r,\theta) & \text{ on } & \Gamma_D = \partial \Omega' \backslash \Gamma_N.
\end{array} \end{equation}
This formulation ensures that the slit in the domain aligns with the exterior of $\Omega'$.

For this problem, the LSGD layer solves the linear system corresponding to the weak formulation of Problem \ref{var:slit}, which by the same process as described for Problem \ref{var:1d}, yields the linear problem of solving $Ac = b$, where
\begin{equation*} \begin{split}
A &= \int_{\Omega'} D\Phi D\Phi^T + \beta \int_{\Gamma_D} \Phi \Phi^T, \qquad
b = \beta \int_{\Gamma_D} g \Phi.
\end{split} \end{equation*}
Note that the matrix $A$ can be significantly smaller than the linear system involved in other scientific computing methods e.g. the finite element method. For P1 finite elements, the linear system would be of size $d_P \left(N_{\text{splines}}\right)^d \times d_P \left( N_{\text{splines}}\right)^d$, whereas in our case $A$ is only of size $d_P N_{\text{cells}} \times d_P \times N_{\text{cells}}$. This reduction in size (and in the cost to solve such problems) arises since we fit polynomials on each partition and not upon each B1-spline basis function. This highlights that our approach provides a nonlinear construction of a reduced finite element space providing an optimal representation of the solution.

The integrals corresponding to $A$ can be calculated via closed-form expressions, as before. However, since $g$ is not a polynomial, we cannot expect exact integration for the integral that defines the vector $b$, and we either can project $g$ into the correct polynomial space, or over-integrate with quadrature of a higher degree to maintain sufficient accuracy.

We compare our solution to results of solving this variational problem using P1 finite elements on three different meshes: two uniform tetrahedral meshes, and an adaptive tetrahedral mesh. The first uniform mesh (FEM U3) is a $3 \times 3$ mesh, chosen so that there are 16 degrees of freedom and $18$ elements, which most closely matches our polynomial-spline construction by fitting linear functions on $N_\text{cells}=16$ cells. The second uniform mesh (FEM U6) is $6 \times 6$, which is the closest match to the size of the linear system that our LSGD layer solves. The adaptive mesh (FEM A) is constructed by an adaptive solver, starting with the $3 \times 3$ uniform mesh and proceeding to adaptively refine the mesh until the number of degrees of freedom is closest to the size of our linear system.
Near the singularity at $r=0$, the optimal adaptive mesh's cells should shrink at a rate of approximately $\sqrt{r}$, where $r$ is their distance to the origin \cite{fenicsbook}. Finite element comparisons are computed using FEniCS \cite{fenics-software}; for more information on how these comparisons are computed, please refer to the supplementary material.

\begin{figure*}[h!]
\centering
\begin{subfigure}[c]{0.24\textwidth}
\centering
\includegraphics[height=1.4in]{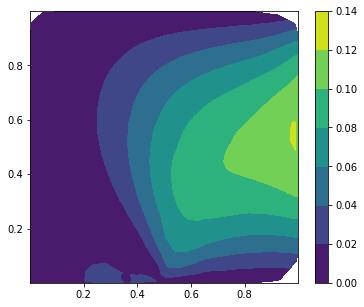}
\caption{Poly-Spline Network}
\end{subfigure}
\begin{subfigure}[c]{0.24\textwidth}
\centering
\includegraphics[height=1.4in]{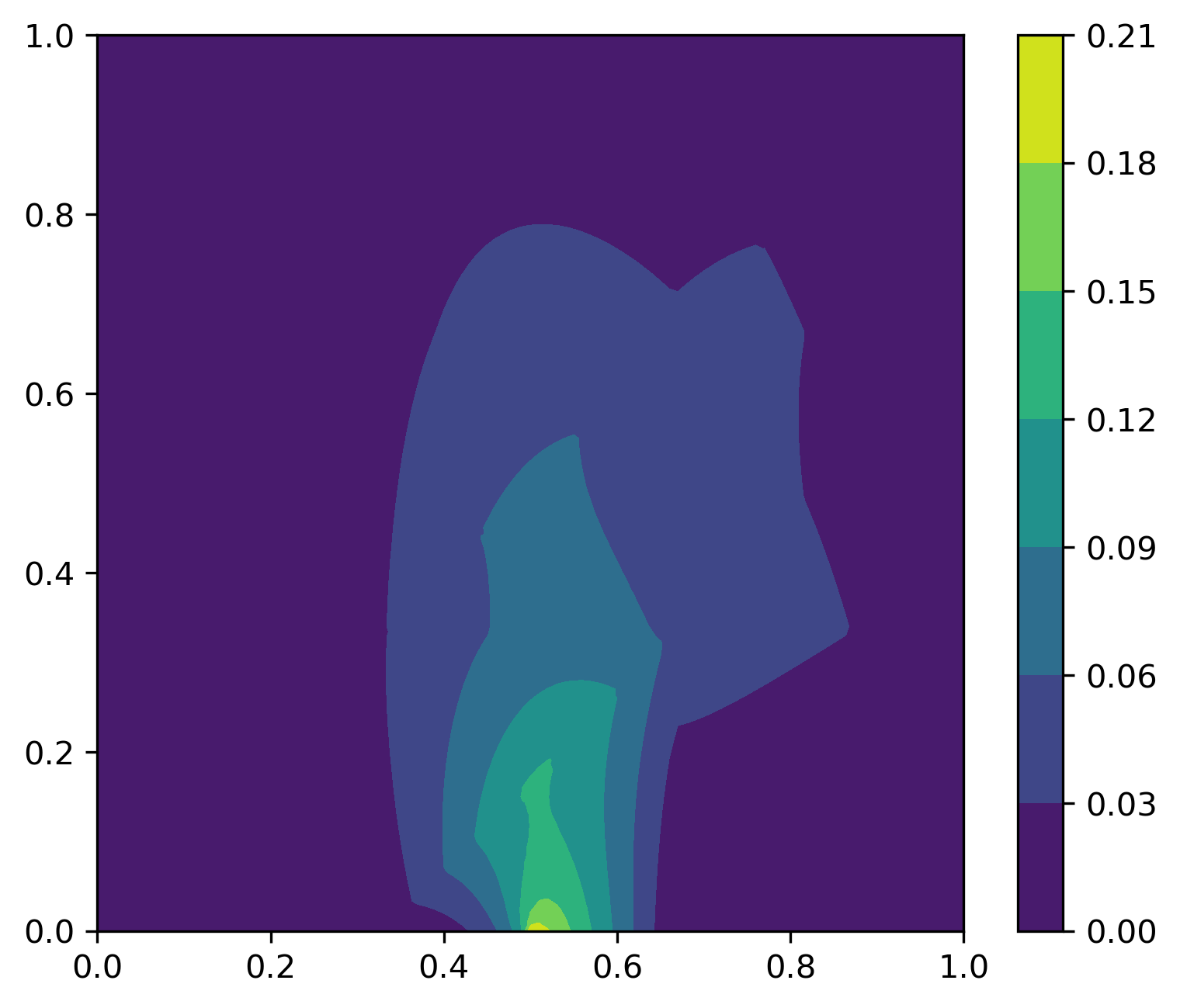}
\caption{FEM, uniform mesh (U3)}
\end{subfigure}
\begin{subfigure}[c]{0.24\textwidth}
\centering
\includegraphics[height=1.4in]{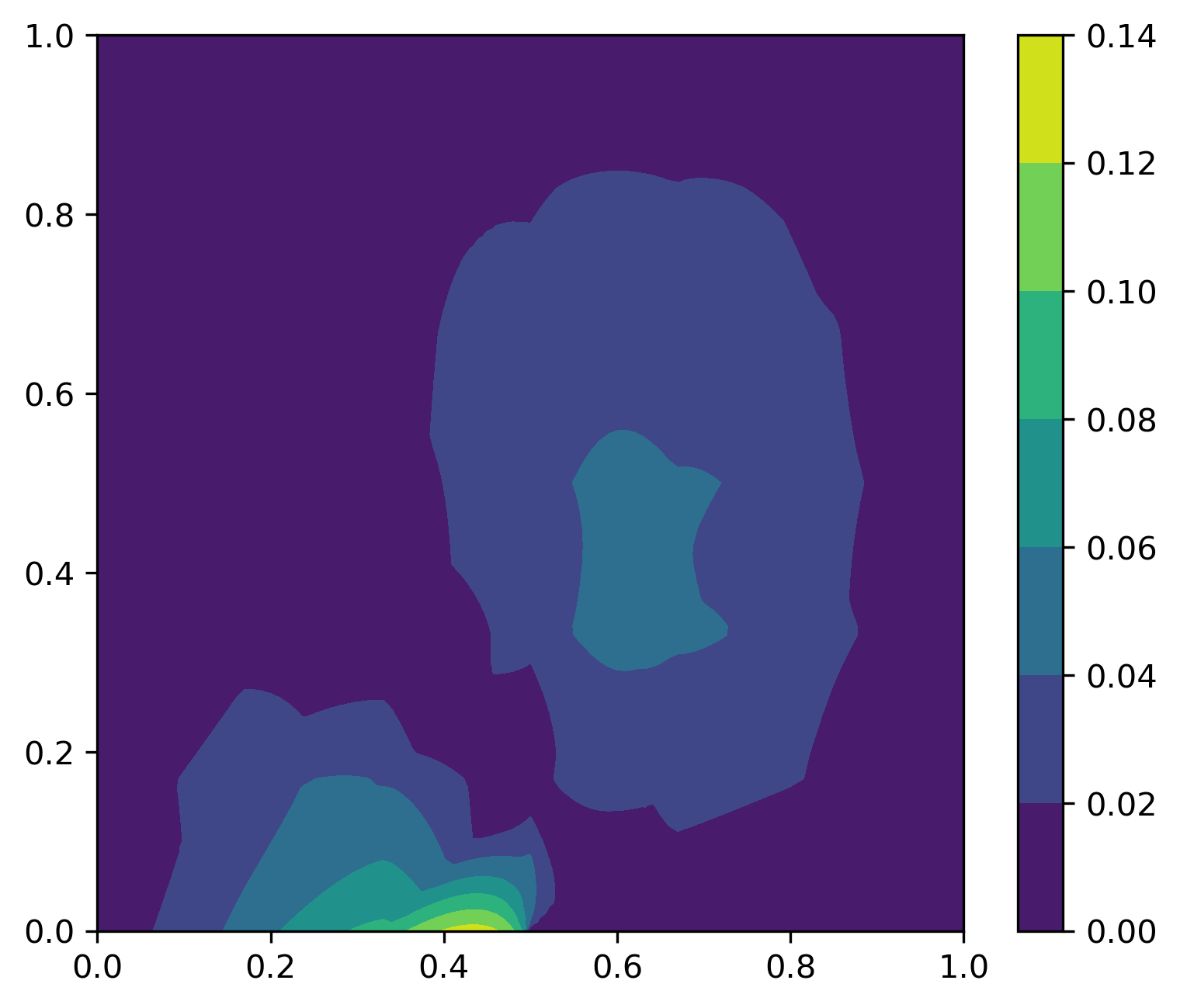}
\caption{FEM, uniform mesh (U6)}
\end{subfigure}
\begin{subfigure}[c]{0.24\textwidth}
\centering
\includegraphics[height=1.4in]{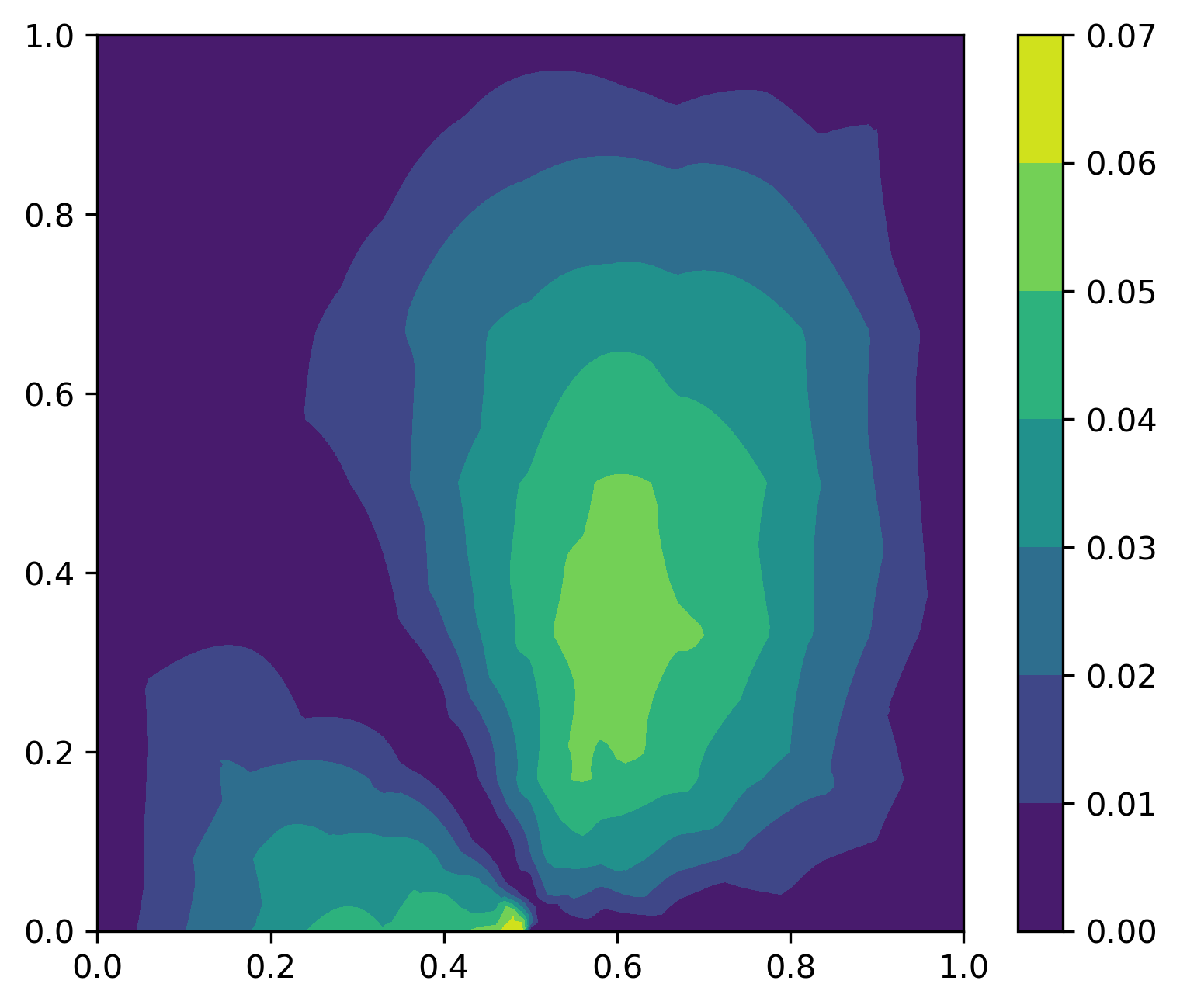}
\caption{FEM, adaptive mesh (A)}
\end{subfigure}
\caption{Pointwise error of our model vs. finite element approximations on various meshes. Note the difference in scale. \label{fig:var2-err}}
\end{figure*}
The results of our method are shown in Figure \ref{fig:var2-err}, with the $L_2$ error listed in Table \ref{table:slit-methods}. We outperform the finite element solver on all of the comparable meshes. 
The plots in Figure \ref{fig:var2-err} re-scale and translate our domain of interest $\Omega'$ from $[-1,1]\times[0,1]$  to $[0,1]^2$, with the slit occurring along the line segment $[0.5,1]\times\{0\}$ along the $x$-axis. 
\begin{table*}[h!]
\centering
\begin{tabular}{llccc}
\hline
\textbf{Method} & \textbf{Mesh Type} & \textbf{\# Cells} & \textbf{Solve Size} & \textbf{$L_2$ Error} \\
\hline
Poly-Spline Network & Adaptive & $16$ & $48 \times 48$ & 0.0262 \\
FEM U3 & Uniform & $18$ & $16 \times 16$ & 0.0362 \\
FEM U6 & Uniform & $72$ & $49 \times 49$ & 0.0231 \\
FEM A & Adaptive & $120$ & $72 \times 72$ & 0.0242 \\
\hline
\end{tabular}
\caption{Comparison of our network vs. adaptive and uniform finite element solutions for Problem \ref{var:slit}. \label{table:slit-methods}}
\end{table*}

The errors that arise in our solution accumulate near the boundary, whereas in the finite element approximations, the errors lie in the interior and near the singularity. This is because our Ritz loss enforcing the boundary condition via penalty, while the finite element method enforces Dirichlet conditions variationally. Still, $L_2$ error of our solution is lower than comparable finite element method solutions, though one could obtain better results working with a boundary conforming Galerkin framework.

\section{Discussion}
Our polynomial-spline network compares favorably to traditional approximation methods, preserving the convergence rates of spline interpolation and polynomial regression. Our ability to learn the spline knots and perform free-knot spline interpolation, like other deep learning methods, allows for adaptivity similar to adaptive mesh refinement methods. However, we obtain a reduced partition of space from the overparameterized fine B-spline grid allowing one to work with polynomials on each adaptively coarsened partition of unity; in this sense we obtain reduced-order model for the adaptive B-spline basis. As a result, the size of the linear system involved in the LSGD optimization step is smaller than if we were to depend on the spline basis functions directly. This reduction is particularly effective when $d > 1$, to avoid the curse of dimensionality. Additionally, the number of parameters in our model does not depend on the degree of the polynomial approximation, since the coefficient values for $c_{\alpha \beta}$ are tabulated via LSGD. We thus obtain an hp-convergent variational framework for solving PDEs that circumvents variational crimes due to inexact quadrature.

\section{Acknowledgments}
\label{sec:acknowledge}
N.~Trask and J.~Actor acknowledge funding under the DOE ASCR PhILMS center (Grant number DE-SC001924) and the DOE Early Career program. A.~Huang acknowledges function under the Laboratory Directed Research and Development program at Sandia National Laboratories. Sandia National Laboratories is a multi-mission laboratory managed and operated by National Technology and Engineering Solutions of Sandia, LLC., a wholly owned subsidiary of Honeywell International, Inc., for the U.S. Department of Energy’s National Nuclear Security Administration under contract DE-NA0003525.

\bibliography{aaai22}

\newpage
\appendix
\section{TECHNICAL APPENDIX}
Presented here is supplementary technical information. Please see the original paper for the main presentation of our results. Videos showing the adaptivity and evolution of the POU cells and spline knots during training for Problem 3 and Problem 4 are included as part of the multimedia supplementary appendix.

\section{Training Parameters}

Training parameters and hyperparameters for each of the four problems are described below.

\subsection{Problem 1}
We aim to recover via regression the function $$f(x) = \sin( 2 \pi x) \qquad x \in [0,1].$$

We employ a dataset of 1000 points, sampled from $[0,1]$ using a random uniform distribution. We construct a separate validation dataset (also used for plotting) of another 1000 points, independently randomly uniformly sampled from $[0,1]$. 

Random NumPy calls are seeded by using a NumPy random number generator with \texttt{seed=1234}. TensorFlow randomness is set via a global random seed of \texttt{seed=1234}, which is independent of the NumPy generator and is restarted each time a new network is trained.

Our polynomial-spline networks are constructed with spatial dimension $d=1$, polynomial degrees $B \in \{0,1,\dots,6\}$, and number of splines in the spline layer $N_\text{splines} \in \{4,8,16,32,64\}$. For each value of $N_\text{splines}$, we test with the number of POU cells $N_\text{cells}=\{1, 2, 4, 8, \dots, N_\text{splines}\}.$ Our model is constructed with a LSGD layer which uses an $L_2$ regularizer of $10^{-10}$ as part of the least-squares solve, to protect against numerical instability in TensorFlow's backwards differentiation of the least squares function call.

During training, we use the Adam optimizer \cite{adam} with a learning rate of $5\times 10^{-3}$, with a batch size of 1000, i.e. performing gradient descent, not stochastic gradient descent. We train for 500 epochs, minimizing mean squared error as our loss function.

\subsection{Problem 2}
We aim to recover via regression the function $$f(x) = \lvert \sin( 3 \pi x^2) \rvert + \lvert \cos(5 \pi x^2) \rvert \qquad x \in [0,1].$$

Our polynomial-spline networks are constructed with spatial dimension $d=1$, polynomial degrees $B \in \{0,1,2,3\}$, and number of splines in the spline layer $N_\text{splines} \in \{4,8,16,32,64\}$. For each value of $N_\text{splines}$, we test with the number of POU cells $N_\text{cells}=\{ 2, 4, 8, \dots, N_\text{splines}\}.$ Our model is constructed with a LSGD layer which uses an $L_2$ regularizer of $10^{-12}$ as part of the least-squares solve, to protect against numerical instability in TensorFlow's backwards differentiation of the least squares function call.

All other parameter and hyperparameter choices are identical to those listed in Problem 1.

\subsection{Problem 3}

We aim to solve the boundary-value partial differential equation
\begin{equation}
\begin{array}{rll}
- d^2u = 2 & \text{ on } & \Omega = [0,1] \\
u = 0 & \text{ at } & \partial \Omega = \{0,1\}.
\end{array} \end{equation}

To do so, we minimize the Euler-Lagrange loss
\begin{equation} 
L(u) = \int_\Omega \frac{1}{2} \left \lVert \nabla u \right \rVert^2 dx + \beta \left( u(0)^2 + u(1)^2 \right).
\end{equation}

Random seeding for TensorFlow initialization is set the same way as in the first two problems. We use the Adam optimizer, enhanced with LSGD, enforced as a layer in the TensorFlow graph, with a $L_2$ regularizer of $10^{-8}$. The penalty parameter $\beta$ in the Euler-Lagrange loss is set to $\beta = 1000$.

For $B=0$, we use a polynomial-spline network with $N_\text{splines} \in \{4,8,\dots, 64\}$ and with fixed $N_\text{cells} =3$. We train for 1000 epochs using the Adam optimizer with an initial learning rate of $0.01$, reducing to $0.005$ halfway through training.
For $B=1$, we use a polynomial-spline network with $N_\text{splines}=5$ and $N_\text{cells}=3$. we train for only 100 epochs, since we recover the true solution faster as it lies in our solution space, and we only use the reduced learning rate of $0.005$.

\subsection{Problem 4}

We aim to solve the problem
\begin{equation}
\begin{array}{rllll}
- \Delta u & = 0 & \text{ on } & \Omega = [-1,1] \times [0,1] \\
\partial_n u & = 0 & \text{ on } & \Gamma_N = [-1,0] \times \{0\} \\
u  &= g(r,\theta) & \text{ on } & \Gamma_D = \partial \Omega \backslash \Gamma_N.
\end{array} \end{equation}

To do so, we minimize the Euler-Lagrange loss
\begin{equation}
L(u) = \int_\Omega \frac{1}{2} \left \lVert \nabla u \right \rVert^2 dx + \beta \int_{\Gamma} (u - g(r,\theta))^2 ds.
\label{eq:EL-var2}
\end{equation}

Random seeding for TensorFlow initialization is set the same way as in the first two problems. We use the Adam optimizer, enhanced with LSGD, enforced as a layer in the TensorFlow graph, with a $L_2$ regularizer of $10^{-8}$. The penalty parameter $\beta$ in the Euler-Lagrange loss is set to $\beta = 1000$.

We use a polynomial-spline network with $B=1$, with $N_\text{splines}=9$ for both the x-axis splines and y-axis splines (recall our spline basis for $d > 1$ is formed via a tensor product of 1D splines), and with fixed $N_\text{cells} = 16$. We train for 3000 epochs using the Adam optimizer with an initial learning rate of $0.01$, reducing to $0.005$ halfway through training. Additionally, we re-initialize the optimizer after every 500 epochs, as there was a noticeable improvement in training when doing so.

We plot using 1600 points randomly uniformly sampled from $[0,1]^2$. These points are not used during training. Contour plots are generated using the PyPlot package's \texttt{tricontourf} function, using default parameters for setting the contours.

\section{Notes on LSGD Implementation}

We expedite the training of our models by using least-squares gradient descent (LSGD) methods \cite{lsgd} as part of our optimization strategy. Following the approach from \cite{lsgd}, we define the function $\Phi: \mathbb{R}^d \rightarrow \mathbb{R}^{N_\text{cells} d_P}$ as $\Phi: x \mapsto \left( \sum_\gamma w_{\alpha,\gamma} \phi_\gamma(x) \right) p_\beta(x)$, and rewrite our model as $$ y(x) = c^T \Phi(x)$$ for a vector of coefficients $c \in \mathbb{R}^{N_\text{cells} d_P}.$ LSGD adds a least-squares solve for $c$ between each gradient step of the first-order optimizer. 

At this point, we have two options for how to insert this least-squares solve into our training routines. Following \cite{lsgd}, we can incorporate this least-squares solve as a TensorFlow Callback (e.g. via NumPy calls to solve the linear system) as a set of operations independent of updating the TensorFlow graph. Alternatively, we can embed the least-squares solve directly into the TensorFlow graph as a TensorFlow Layer, thereby directly enforcing that all outputs of the network lie on the manifold of best-fit solutions regarding the coefficients of the outermost layer.

\begin{figure*}[ht]
\centering
\begin{subfigure}{0.9\columnwidth}
\centering
\includegraphics[width=0.9\columnwidth]{hp-approx-layer}
\caption{TensorFlow Layer LSGD implementation}
\end{subfigure}
\begin{subfigure}{0.9\columnwidth}
\centering
\includegraphics[width=0.9\columnwidth]{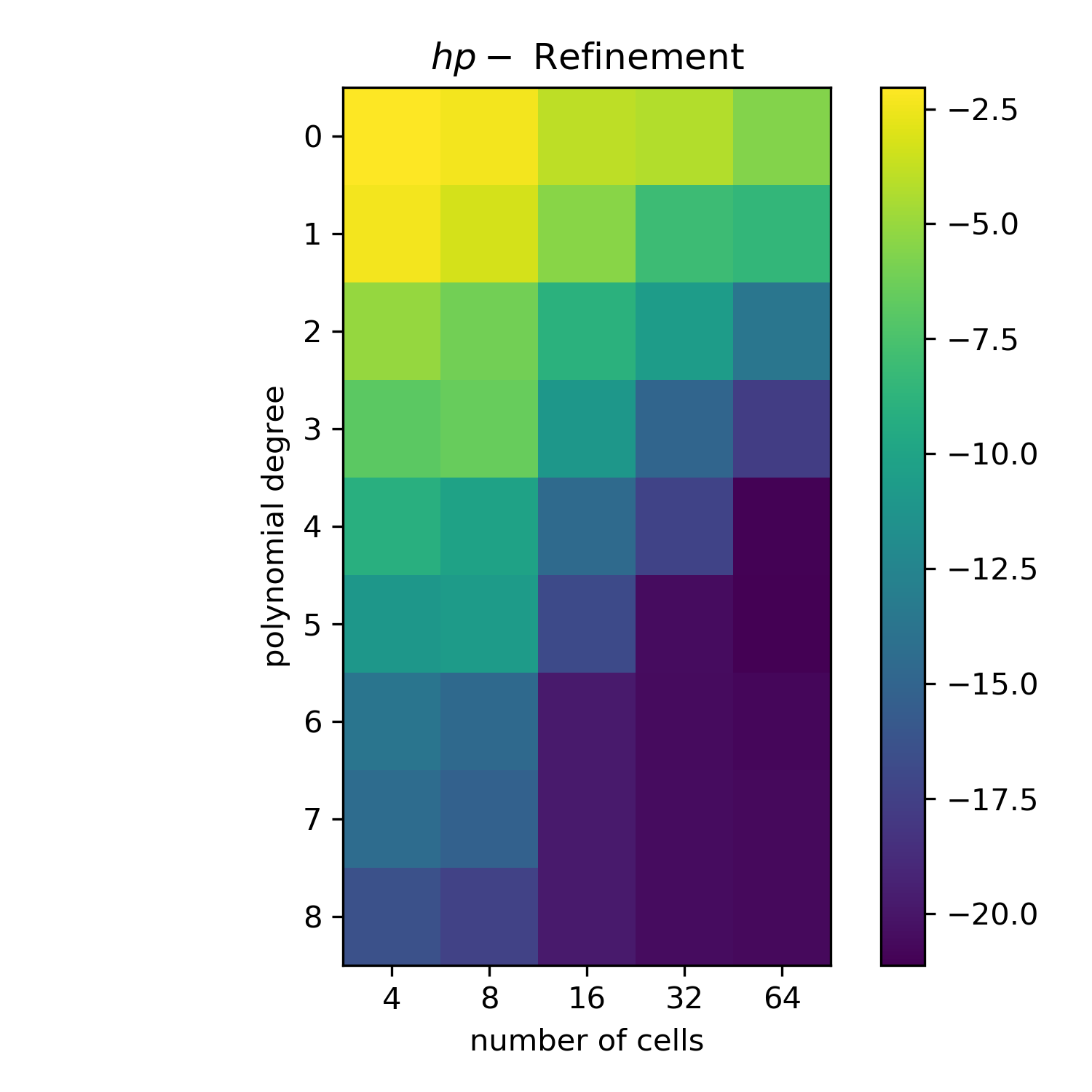}
\caption{TensorFlow Callback LSGD implementation}
\end{subfigure}
\caption{Comparison of $hp-$ convergence results using TensorFlow Callback vs. TensorFlow Layer implementations of LSGD. \label{fig:lsgd-comparison}. }
\end{figure*}
Note that there are practical differences in using the LSGD callback implementation vs. adding the least-squares step as a layer to the TensorFlow graph. Using the callback implementation of LSGD enables accuracy up to machine precision, but in practice training takes more iterations until convergence, since our iterates after each full training step no longer live on the manifold of least-squares solutions with respect to the basis constructed by the inner layers. In contrast, the layer implementation directly enforces that the output of the network lies on this manifold, and as a result training  generally converges faster in practice, but the accuracy of the computation is limited by the stability of the backwards-differentiability of the least-squares solver in e.g. TensorFlow, which is substantially less accurate than machine precision. In our experiments, we could only achieve $O(10^{-8})$ accuracy with the LSGD layer, as compared to $O(10^{-20})$ or better with the callback. To demonstrate, we repeat the $hp-$ convergence test for Problem 1; the results of using the TensorFlow Callback vs. TensorFlow Layer implementations are seen in Figure \ref{fig:lsgd-comparison}.

For code implementations of both the TensorFlow Layer and TensorFlow Callback implementations, please see the supplementary code appendix.

\section{Notes on Spline Layer Implementation}

To build our polynomial-spline network, we build B1-spline basis functions for our domain $\Omega \subset \mathbb{R}^d$ as a tensor product of B1-spline basis functions along each dimension. For each dimension, we construct a B1-spline layer, whose knots are parameterized to accommodate TensorFlow backwards differentiation during training. 

We outline our implementation of 1-dimensional B1-spline basis functions on the interval $[0,1]$. These basis functions are fully described by their knots $\{0=t_0, t_1,\dots, t_{N-1}, t_N = 1\}$. These knots must remain ordered, else the basis functions will no longer form a POU, nor will they necessarily remain a basis.

In what follows, let $\sigma: \mathbb{R}^N \rightarrow [0,1]^N$ be the softmax function. We parameterize a B1-spline layer in TensorFlow with a vector $\mu \in \mathbb{R}^N$, where for $i=1,\dots, N$, the values $\sigma(\mu)_i$ define the interval between knots $t_{i-1}$ and $t_{1}$.
Additionally, define the matrix $A \in \mathbb{R}^{N \times n}$ as the lower-triangular matrix of all $1$'s, that is, $$A_{ij} = 1 \quad \text{if} \quad i \ge j.$$
From the vector $\mu$ and the matrix $A$ we can then recover the vector of knots via the expression
$$t_0 = 0; \qquad\qquad t_{1:N} = A \sigma(\mu).$$

This method for expressing the knots has several advantages. First, we ensure that our knots remain ordered: by definition, 
\begin{equation*} \begin{split}
t_{i+1} &= \left( A \sigma(\mu) \right)_{i+1} \\
&= \sum_{j=1}^{i+1} \sigma(\mu)_j \\
&= \sum_{j=1}^i \sigma(\mu)_j + \sigma(\mu)_{i+1} \\
&= t_i + \sigma(\mu)_{i+1} \\
&> t_i,
\end{split} \end{equation*}
where the last inequality is strict due to the softmax function always being positive. Second, by the same reasoning, the interval widths always remain positive as well, ensuring that our B1-spline basis functions never truly collapse to zero. As a result, we can always define quadrature points within the support of this hat function, ensuring the stability of our integration routines.
Third, we can enforce convex combinations so that our knots span our domain, without needing to enforce inequality or even equality constraints, performing unconstrained optimization during training instead of constrained optimization to enforce that our splines remain valid, even as the spline knots evolve during training. 

To see how the splines evolve while the knots remain valid, see the multimedia supplemental submissions. In \texttt{slit\_domain\_resampled.gif}, the solution to Problem 4 is plotted after each epoch (resampled for one frame every 10 epochs), and along each axis the corresponding splines that form the B1-spline tensor product basis are plotted as well. As training continues, we see the splines along the $x$-axis slowly being pulled towards the singularity at the point $(0.5,0)$, which is where the slit domain ends (after our remapping for $\Omega$). Similarly, in \texttt{poisson1d\_training\_resampled.gif}, the solution to Problem 3 is displayed at each epoch; looking at the evolution of the intervals in the derivative plots, we see the mesh first adapting to resolve the boundary conditions, pushing the intervals towards the endpoints $x=0$ and $x=1$, and then later during training pulling intervals back towards the center, becoming nearly uniformly spaced by the end of training.

For code that implements this layer, please see the supplementary code appendix.

\section{Supplementary Results: Problem 3,  $B\ge1$}

We repeat our experiments from before using $B = 1$, since in the case $B\ge 1$, the true solution to Problem 3 lies in our solution space. Training details are described above.

\begin{figure*}[ht!]
\centering
\includegraphics[width=0.9\textwidth]{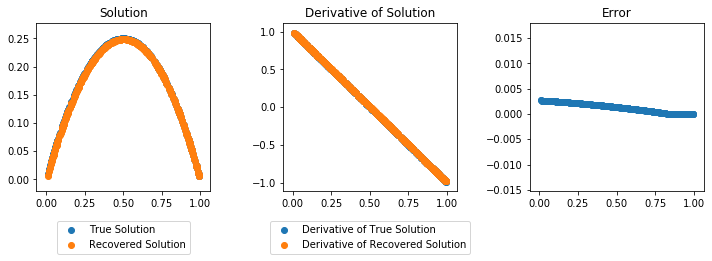}
\caption{Results for $B=1$ for Problem 3, which should recover our solution exactly. We see that some error remains, due to enforcing the boundary conditions via penalty and due to the stability of the automatic differentiation of the linear solve in LSGD. \label{fig:p3-b1}}
\end{figure*}
After training for only 100 epochs, we recover an $L_2$ error of $0.009$, which is better than the results obtained by training our $B=0$ models for significantly more epochs. Results are shown in Figure \ref{fig:p3-b1}. In the error plot on the right, we see that there is error is dominated by enforcing the boundary conditions, since these are enforced by penalty instead of variationally (as a Finite Element method would do). Even though we are not using mean squared error as our loss, we recover an MSE of $2.7252 \times 10^{-6}$, which is close to our observed limitations from the stability of the automatic differentiation of the the Cholesky decomposition in the linear solve as part of the LSGD implementation.

\section{FEM Comparisons}
Our finite element comparisons for Problem 4 are implemented via FEniCS. Our mesh generation for the non-adaptive problems, and the starting mesh for the adaptive problem, are uniform triangular meshes, created via FEniCS's function \texttt{UniformSquareMesh}. We employ P1 Lagrange finite elements as our approximation space upon the constructed meshes. Our errors are computed by overintegrating with a quadrature 3 degrees higher than necessary.

For the non-adaptive comparisons, we use the default values for the linear solver using the \texttt{solve} function. For the adaptive problem, we use a mesh energy function $J = \int_\Omega u^2 dx$ to indicate where refinement is necessary. Adaptivity is then computed using the built-in \texttt{AdaptiveLinearVariationalSolver} function, using default parameters except for the linear dual variational solver for computing the error control function, which is set to use the conjugate gradient method. We perform adaptive mesh refinement until we reach a solver tolerance level of 0.002, which is the closest step to where the problem size on the refined mesh is similar to that of the linear solve in the polynomial-spline network.

We plot the finite element comparisons on a uniform triangular mesh of size $100 \times 100$, again with the mesh generated by the FEniCS function \texttt{UniformSquareMesh}.

Code for solving these variational problems via finite elements is included in the supplementary material.

\end{document}